\documentclass[runningheads]{llncs}
\usepackage{graphicx}
\usepackage{subcaption}
\usepackage{float}
\usepackage{orcidlink}
\hypersetup{hidelinks}

\usepackage{xcolor}
\newcommand\js[1]{{#1}} 
\newcommand\yh[1]{{#1}} 
\newcommand\hw[1]{{#1}} 
\newcommand\ik[1]{{#1}} 

\begin{document}

\title{SSASS: Semi-Supervised Approach for Stenosis Segmentation}
%
\author{
In Kyu Lee\inst{1}\orcidlink{0000-0001-5554-808X} \and
Junsup Shin\inst{1}\orcidlink{0000-0003-3280-1622} \and
Yong-Hee Lee\inst{1}\orcidlink{0000-0001-6047-701X} \and
Jonghoe Ku\inst{1}\orcidlink{0009-0003-1260-619X} \and
Hyun-Woo Kim\inst{1,*}\orcidlink{0009-0003-2740-0397}
}
\authorrunning{I.K. Lee et al.}
%
\institute{
Medipixel Inc, Seoul, Republic of Korea\\
\email{\inst{*}hyunwoo.kim@medipixel.io}\\
}
\maketitle              

\begin{abstract}

\hw{

Coronary artery stenosis is a critical health risk, and its precise identification in Coronary Angiography (CAG) can significantly aid medical practitioners in accurately evaluating the severity of a patient's condition. 
The complexity of coronary artery structures combined with the inherent noise in X-ray images poses a considerable challenge to this task. 
To tackle these obstacles, we introduce a semi-supervised approach for cardiovascular stenosis segmentation. 
Our strategy begins with data augmentation, specifically tailored to replicate the structural characteristics of coronary arteries. 
We then apply a pseudo-label-based semi-supervised learning technique that leverages the data generated through our augmentation process. Impressively, our approach demonstrated an exceptional performance in the Automatic Region-based Coronary Artery Disease diagnostics using x-ray angiography imagEs (ARCADE) Stenosis Detection Algorithm challenge by utilizing a single model instead of relying on an ensemble of multiple models.
This success emphasizes our method's capability and efficiency in providing an automated solution for accurately assessing stenosis severity from medical imaging data.

}

\end{abstract}

\section{Introduction}

\yh{

The heart is one of the most vital organs in the human body, responsible for circulating blood with self-feeding myocardium.
Blood to the myocardium is supplied by the coronary artery, which can be affected by the accumulation of atherosclerotic plaque on its inner walls, called coronary artery disease.
The diseased lumen disturbs the blood flow, which may lead to angina or acute myocardial infarction depending on the severity of the stenosis~\cite{cad_lead_to_16}. 

Coronary artery disease has been a global burden in terms of the leading cause of death worldwide, also imposing healthcare system expenses~\cite{global_burden_19,global_burden_20,global_burden_22}.
Cardiovascular disease caused 27\% of the world's deaths~\cite{who_cvd_rpt}, while 41.2\% of the deaths attributable to cardiovascular disease in the United States were 
accounted for coronary heart disease~\cite{aha_stats_23}.
Consequently, coronary artery disease is responsible for about 10\% of all deaths, highlighting its importance as a prominent subject in the medical domain.

The recommended apparatus for therapeutic decision-making of coronary artery disease is invasive coronary angiography (ICA)~\cite{aha_guideline_23}.
However, the visual assessment of coronary angiograms obtained from ICA often encounters the following complexities: overlap of background structures, low contrast with surrounding tissues, uneven distribution of contrast medium, convoluted vessel morphology, and the inherent difficulty in the interpretation of the projected 3D structure~\cite{coronary_angiography_textbook}.


These complexities make the evaluation of stenosis in coronary angiograms subjective and time-consuming. 
Especially during interventional procedures, without the segmentation of stenosis, only qualitative analysis of stenosis can be performed through visual inspection. 
Such qualitative analysis may lead to inter- and intra-operator variability. 
Therefore, there is a compelling need for automatic segmentation of stenosis to enable accurate quantitative analysis and mitigate these challenges.




}

\begin{figure}[t]
\includegraphics[width=\textwidth]{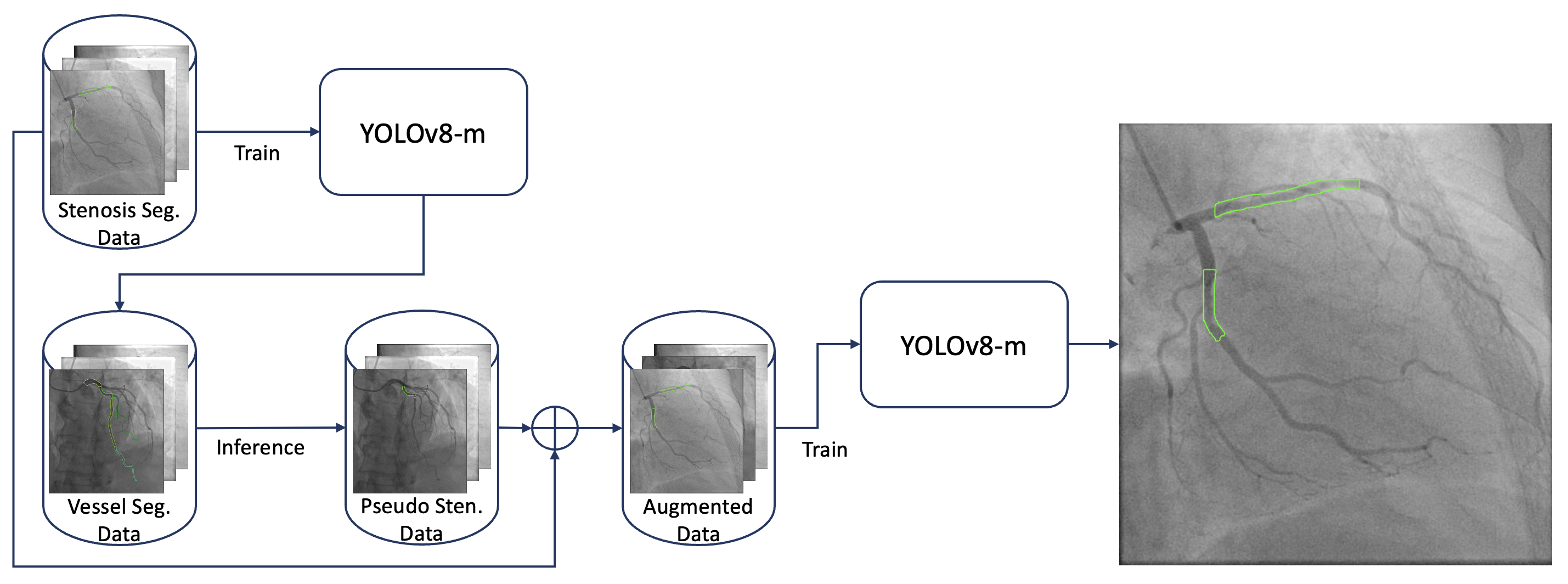}
\caption{An overview of our semi-supervised learning pipeline for stenosis segmentation. YOLOv8-m model is first trained with the stenosis segmentation dataset. The trained model is used to generate pseudo-labels in the vessel segmentation dataset. The final YOLOv8-m model is trained with both the stenosis segmentation dataset and the pseudo-labeled vessel segmentation dataset.} \label{fig1}
\end{figure}

\js{
Recent advancements in deep learning~\cite{imagenet,resnet,unet} have made deep neural networks increasingly viable within clinical fields~\cite{medical_image_processing_survey}. 
However, training a robust deep learning model suitable for clinical applications necessitates tremendous data with annotations. 
The process of annotating medical data demands a significant amount of labor and can only be carried out by highly skilled medical experts.
To address this issue, we propose a straightforward yet accurate pseudo-label-based semi-supervised learning approach~\cite{pseudo_label_ssl}, tailored to the distinctive features of coronary arteries.

Our proposed method presents three compelling attributes that enable effective stenosis segmentation in the challenge:
}

\js{
\begin{itemize}
    \item \textbf{Data Augmentation with Respect to Structural Characteristics of Coronary Arteries}: Thoughtfully designed data augmentation techniques that align with the structural nuances of coronary arteries, enhancing the model's ability to generalize.
    \item \textbf{Pseudo-label-based Semi-supervised Learning}: Leveraging additional angiograms by using the vessel segmentation dataset through pseudo-labeling, thereby augmenting the learning process.
    \item \textbf{No Model Ensemble Usage}: Opting to avoid model ensembles, optimizing for inference speed and memory efficiency, all while achieving the top performance in the challenge.
\end{itemize}
}

\js{
Our augmentation strategy, primarily employing projective transforms~\cite{augmentation_survey}, encompasses affine and perspective transformations. Though seemingly fundamental, these transformations prove to be highly effective in capturing the complexities of the coronary artery's three-dimensional structure~\cite{coronary_angiography_textbook}. 
The subtle shifts and perspectives introduced through these transformations significantly enrich the training dataset. This augmentation approach equips the segmentation model to comprehend not only the complex structure but also the inherent structural variations among individuals.
By keeping our augmentation methodology rooted in projective transformations, we leverage a powerful tool that intuitively aligns with the structure of coronary arteries.
These transformations lay a solid foundation for the model to discern and delineate stenotic regions accurately within angiography images.

Given the non-overlapping angiograms in the coronary artery and stenosis segmentation datasets, we utilized the segmentation dataset as a source of unlabeled images to drive our semi-supervised learning~\cite{dataset,pseudo_label_ssl}. Moreover, we carefully considered the unique structural characteristics of coronary arteries in crafting our data augmentation strategies. This approach led to superior performance compared to relying solely on supervised learning from the stenosis dataset. Remarkably, we refrained from utilizing model ensembles to bolster the inference score, prioritizing both inference speed and optimal performance in the challenge.
}

\js{
Semi-supervised learning~\cite{ssl_survey} strikes a balance between supervised learning~\cite{resnet,efficientnet} that requires abundant labeled data and unsupervised learning~\cite{ul_ssl_survey,autoencoder,gan} that operates with only unlabeled data. In such fields like medical image processing~\cite{unet,medical_image_processing_survey,ref_breast_tumor}, labeling images involves extensive labor of highly trained experts. On the other hand, collecting medical images without any annotation is relatively easy. Semi-supervised learning combines this benefit of supervised and unsupervised learning. By effectively leveraging the unlabeled data in conjunction with the limited labeled data, semi-supervised learning addresses the challenge of data scarcity and enhances the model's performance without a prohibitive increase in annotation costs. The model learns to generalize patterns and features from the labeled data, while also exploiting the information embedded in the unlabeled data for training.

There are various methods of semi-supervised learning, but among them, exploiting the pseudo-label of the unlabeled images is one of the most fundamental approaches~\cite{pseudo_label_ssl}. 
The proposed method initially trains the stenosis segmentation model using the angiograms and the corresponding stenosis annotation. Subsequently, predictions are extracted from the segmentation dataset~\cite{dataset} to be utilized as pseudo-labels. 
Finally, the model is trained with the stenosis dataset and the images from the coronary artery segmentation dataset with the pseudo-label. 
The details of the proposed method are described in Section 2.

}
\section{Method}
\ik{
In this section, we present a comprehensive overview of the ARCADE challenge datasets for stenosis segmentation and coronary artery segmentation. 
In addition, we provide a brief introduction to the evaluation metrics used in this challenge.
We then introduce our proposed methods for generating pseudo-labels and the subsequent training of our deep learning model. 
To provide a holistic understanding, we also delve into the specifics of our implementation details, encompassing essential training hyperparameters and inference hardware specifications.
}

\subsection{ARCADE Challenge}
\subsubsection{Dataset}
\begin{figure}[t]
\includegraphics[width=\textwidth]{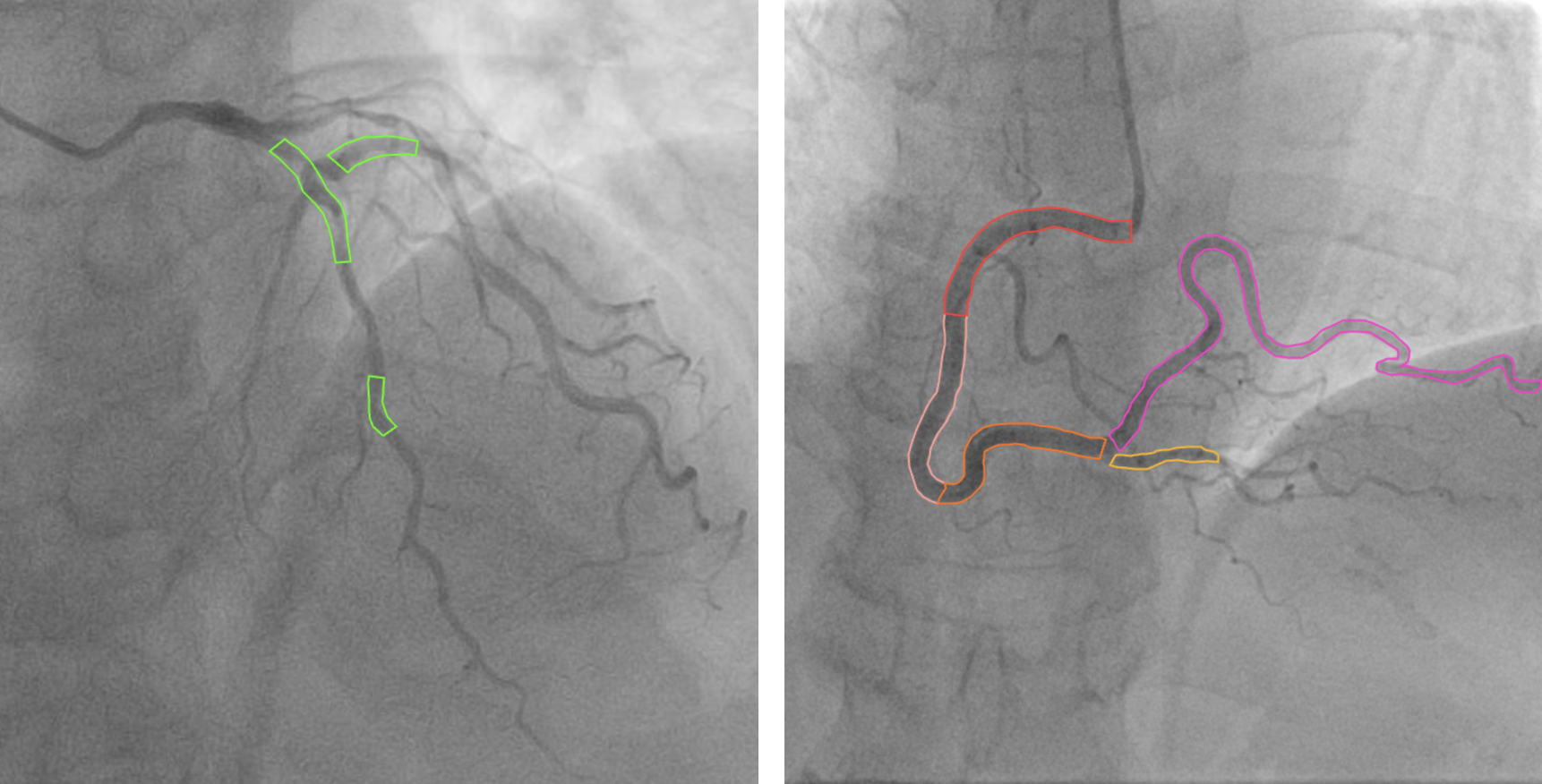}
\caption{Examples of the ARCADE challenge images. Left: An image from the stenosis segmentation dataset, where each stenosis instance is delineated by light green contour lines.
Right: An image from the coronary artery segmentation dataset, where different colors in the contour lines represent distinct segments of the vessel.} \label{fig:sample}
\end{figure}

\ik{
The ARCADE challenge provides two CAG datasets: the coronary artery segmentation dataset and the coronary stenosis segmentation dataset. 
Each dataset is divided into subsets, encompassing 1000 training images with segmentation labels, 200 validation images with segmentation labels, and 300 test images. 

Within the stenosis segmentation dataset, each image is characterized by the presence of at least one stenosis region, and the entirety of stenotic plaques within these images is delineated. 

The coronary artery segmentation dataset adheres to the SYNTAX (SYNergy between PCI with TAXUS\textsuperscript{TM} and Cardiac Surgery) Score definitions~\cite{syntax_origin} for segmentation. 
It is worth noting that explicit stenosis annotations are not provided within this dataset.
The labels for both datasets have been meticulously annotated by medical experts~\cite{dataset}.

For a visual representation of the labeled data, examples of annotated images are shown in Fig. \ref{fig:sample}, illustrating the segmentation tasks for stenosis and coronary artery.
}

\subsubsection{Evaluation}
\ik{
The only evaluation metric used in this challenge was the mean F1 score, also referred to as the Dice coefficient in segmentation tasks. 
The F1 score serves as a measure of the harmonic mean between precision and recall.

\begin{equation}
F1 = 2 \frac{precision\cdot recall}{precision+recall}
\end{equation}

Precision and recall can be calculated using true positives (TP), false positives (FP), and false negatives (FN):

\begin{equation}
precision = \frac{TP}{TP+FP}
\end{equation}

\begin{equation}
recall = \frac{TP}{TP+FN}
\end{equation}

In scenarios where there might be more than one stenosis instance within a single image, F1 scores for all stenosis instances within each image were assessed.
The F1 score for an individual image is determined by averaging the F1 scores of all stenosis instances within that image.
To compute the mean F1 score, an average of F1 scores was taken across all images, represented as:

\begin{equation}
mean F1 = \frac{1}{N} \sum_{i=1}^{N} \frac{1}{M_i} \sum_{j=1}^{M_i}  F1_{ij}
\end{equation}

$M$ represents the total number of stenosis instances, and $N$ denotes the total number of images in the evaluation dataset. 
The mean F1 score provides an aggregated measure of segmentation performance across all stenosis instances and images.

The evaluation process for a single image is strictly constrained with a maximum time limit of 5 seconds.
If the inference time surpasses this predefined threshold, the image receives a score of 0.
In cases where two submissions yield mean F1 scores that exhibit no more than a 0.1\% difference, the winning submission is determined based on the criterion of shorter inference time. 
This rule ensures that the evaluation process maintains fairness and efficiency, encouraging submissions not only to produce accurate results but also to do so within a reasonable time frame.

}

\subsection{Proposed method}

\ik{
\subsubsection{Data Augmentation}
We trained a YOLOv8m model for stenosis instance segmentation. 
Given the nature of CAG image acquisition during interventional procedures, we adapted strong geometric data augmentation, including translation, rotation, and scaling. 
Unlike the coronary artery segmentation task, where location information is critical, we have incorporated both vertical and horizontal flips into our augmentation strategy, further diversifying the dataset to enhance model robustness. 
After converting the original single-channel grayscale images into three-channel images, we have incorporated hue, saturation, and value (HSV) data augmentation techniques. 
More detailed list and parameters of our data augmentation is summarized in Table \ref{tab1}.

\subsubsection{Pseudo-label}
Despite the absence of explicit stenotic region labels in the coronary artery segmentation dataset, it is important to note that stenosis regions are indeed present within the images of this dataset. 
To address this challenge and enhance our stenosis segmentation model, we adopted a semi-supervised learning approach. 
The key components of this methodology are outlined in Fig. \ref{fig1}, which provides a schematic representation of our semi-supervised learning procedure.

Our approach commenced with the training of the YOLOv8m model using the provided stenosis dataset. 
Subsequently, we employed this model for inference on the vessel segmentation dataset. 
For an optimal balance between precision and confidence, we adaptively selected a confidence threshold for the predictions generated on the vessel segmentation dataset. 
We then collected all predictions that surpassed this specified threshold, effectively assembling a pseudo-label dataset.

In the next stage of our methodology, we combined this newly generated pseudo-label dataset with the original stenosis dataset. 
The combined dataset served as the training data for our second stenosis segmentation model, resulting in an improved and more robust model for stenosis instance segmentation.

}

\subsection{Implementation details}

\ik{
\subsubsection{Training procedures}
Two YOLOv8m models were trained as part of our pipeline, each employing identical training hyperparameters. 
The first model underwent training exclusively with the stenosis dataset, while the second model benefited from an augmented dataset comprising both the original stenosis dataset and the pseudo-labeled stenosis dataset. 

Throughout the training process, input images were consistently resized to a resolution of 640x640 pixels and subsequently normalized to ensure uniformity and facilitate model convergence.
For the optimization of both models, we adopted a stochastic gradient descent (SGD) optimizer with a learning rate set at 0.01 and a weight decay of 0.0005. 
To expedite convergence and improve training stability, we used a cosine annealing learning rate scheduler with warm restarts.
Our training objective encompassed optimizing both bounding box and segmentation predictions. For this purpose, we employed the binary cross-entropy loss function.
After training the model for 300 epochs, we selected the model parameters from the epoch that exhibited the lowest validation loss as our submission for evaluation in the final phase of the challenge.

\subsubsection{Evaluation hardware}
The docker container image for final evaluation was uploaded to the grand challenge platform and underwent evaluation on Amazon Web Services (AWS). The evaluation hardware utilized during this process was equipped with an NVIDIA T4 GPU featuring 16GB of memory and 8 CPUs with a total of 30GB of memory.
}

\begin{figure}
    \centering
    \begin{minipage}{0.24\linewidth}
        \includegraphics[width=\linewidth]{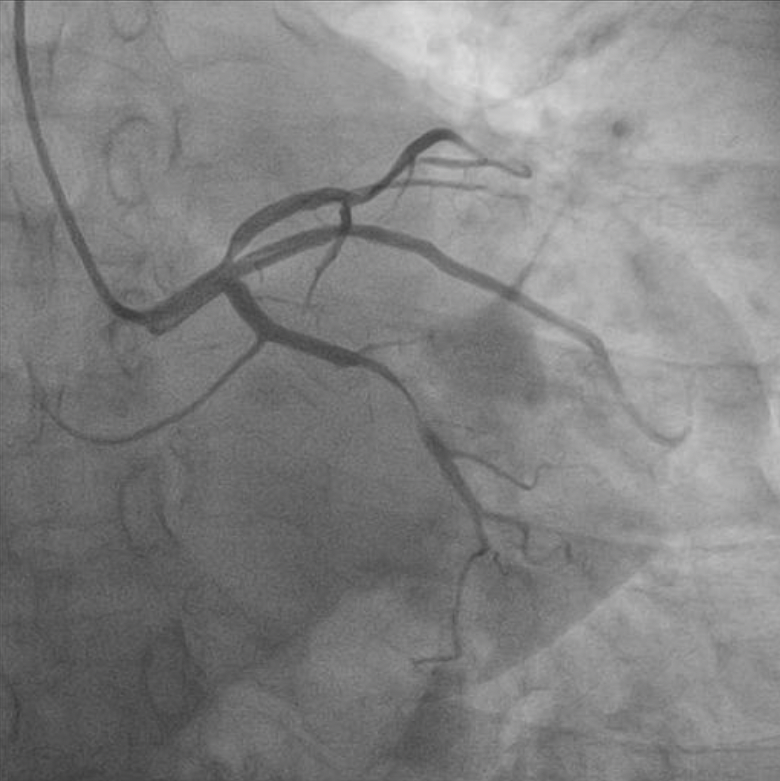}
        \includegraphics[width=\linewidth]{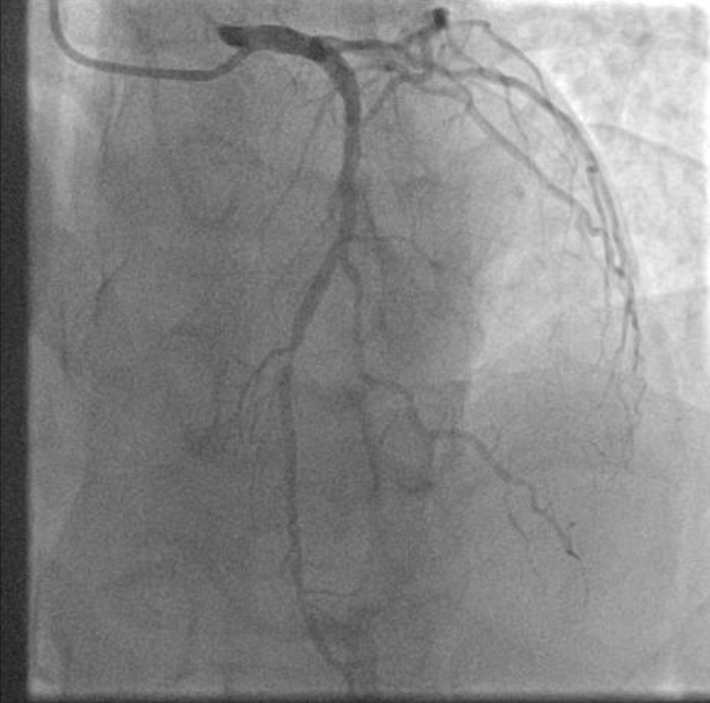}
        \includegraphics[width=\linewidth]{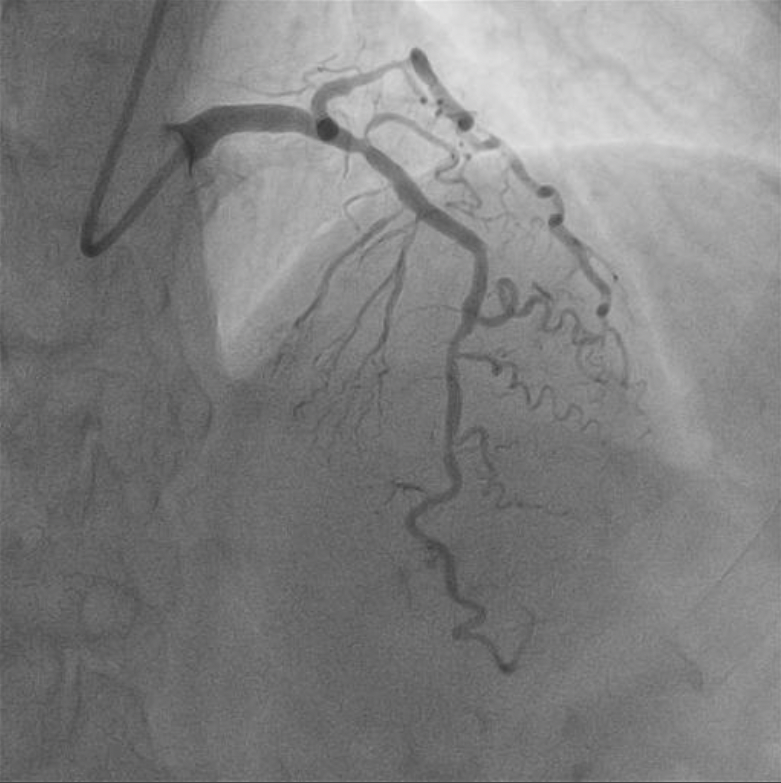}
        \includegraphics[width=\linewidth]{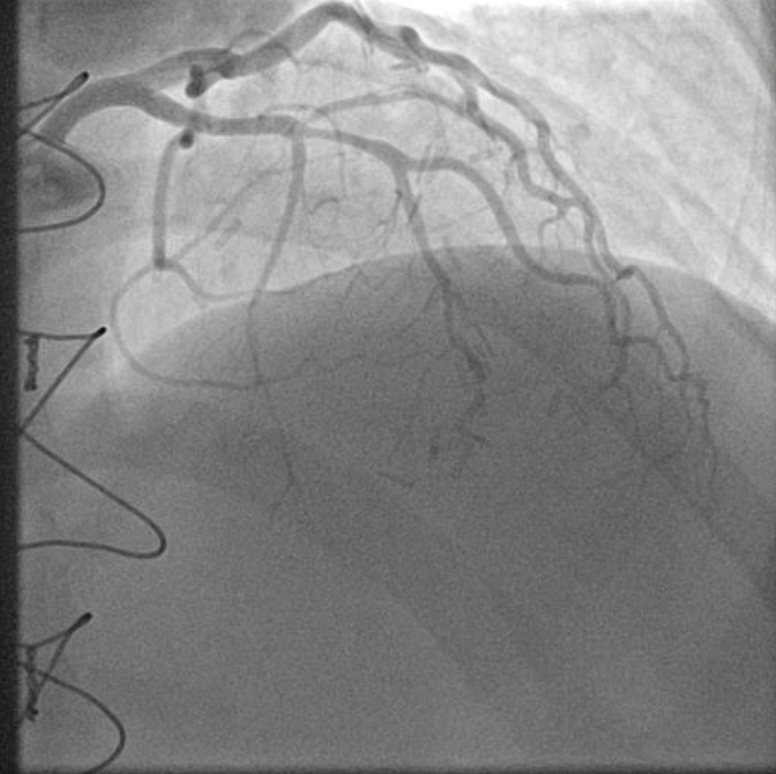}
        \includegraphics[width=\linewidth]{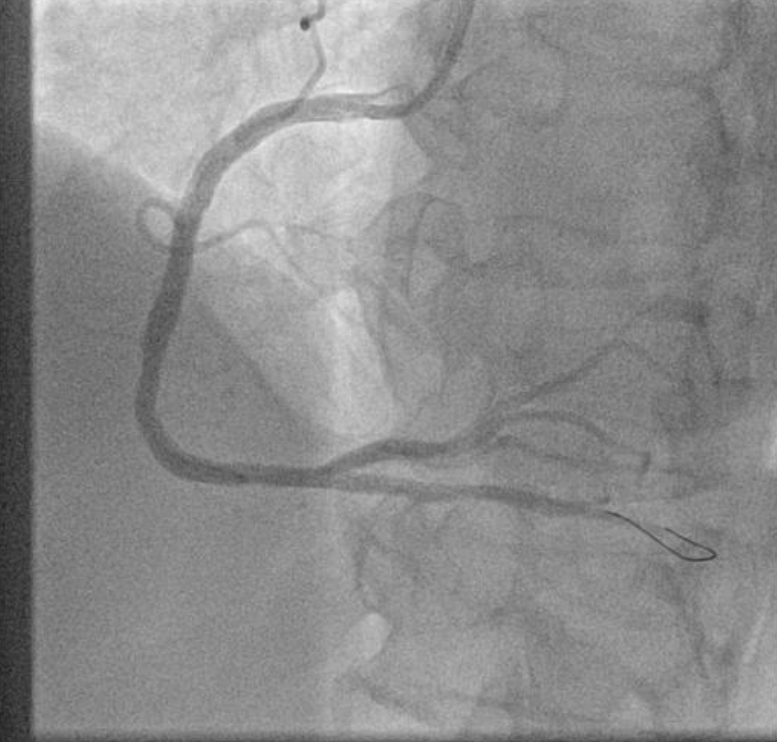}
        \includegraphics[width=\linewidth]{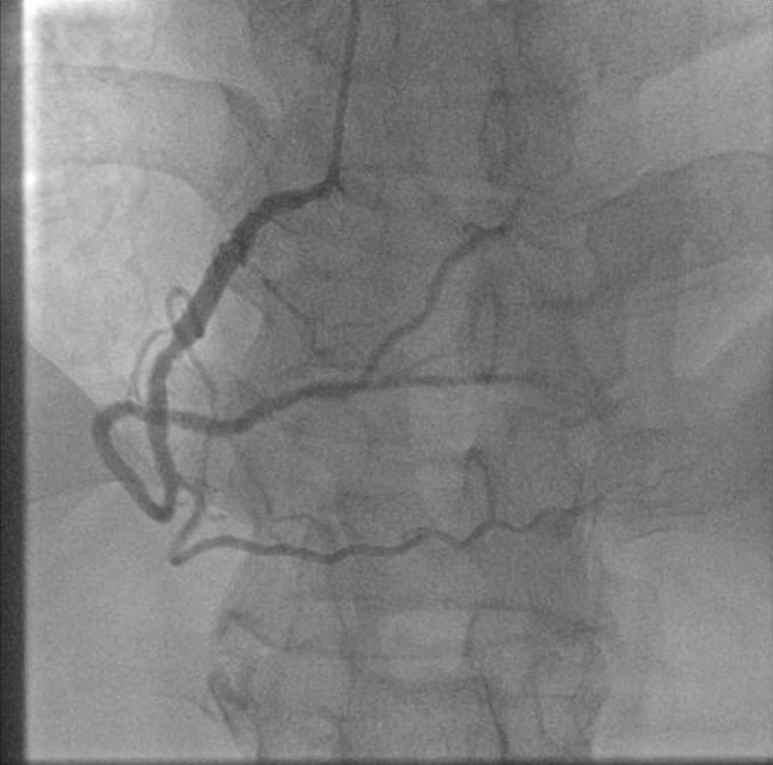}
        \subcaption{CAG}
        \label{fig:qual_img}
    \end{minipage}\hfill
    \begin{minipage}{0.24\linewidth}
        \includegraphics[width=\linewidth]{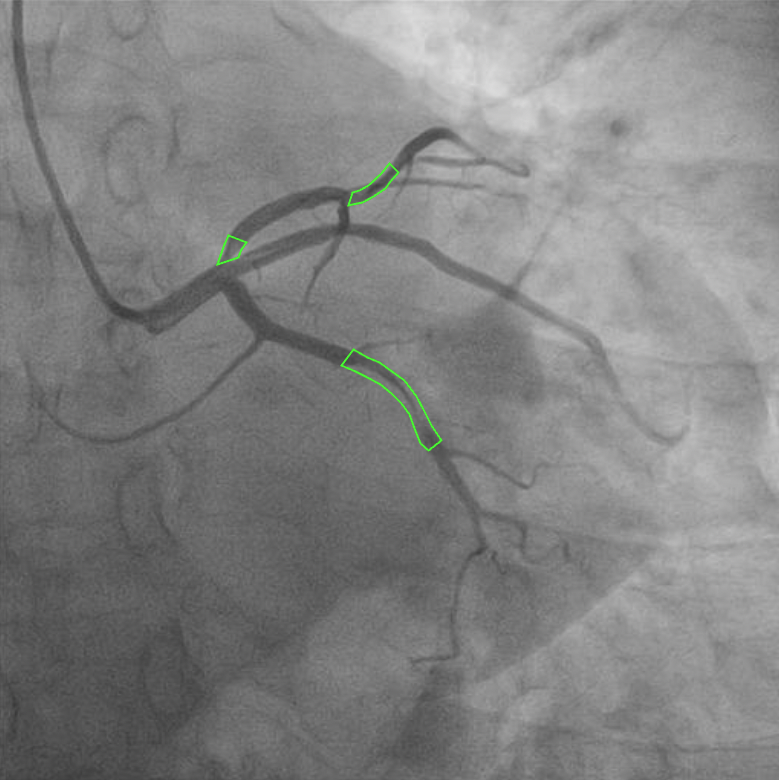}
        \includegraphics[width=\linewidth]{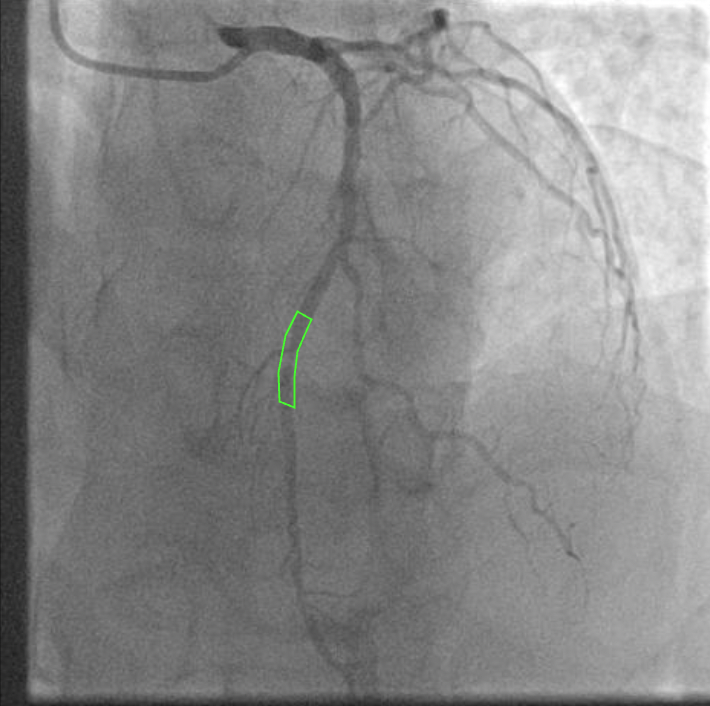}
        \includegraphics[width=\linewidth]{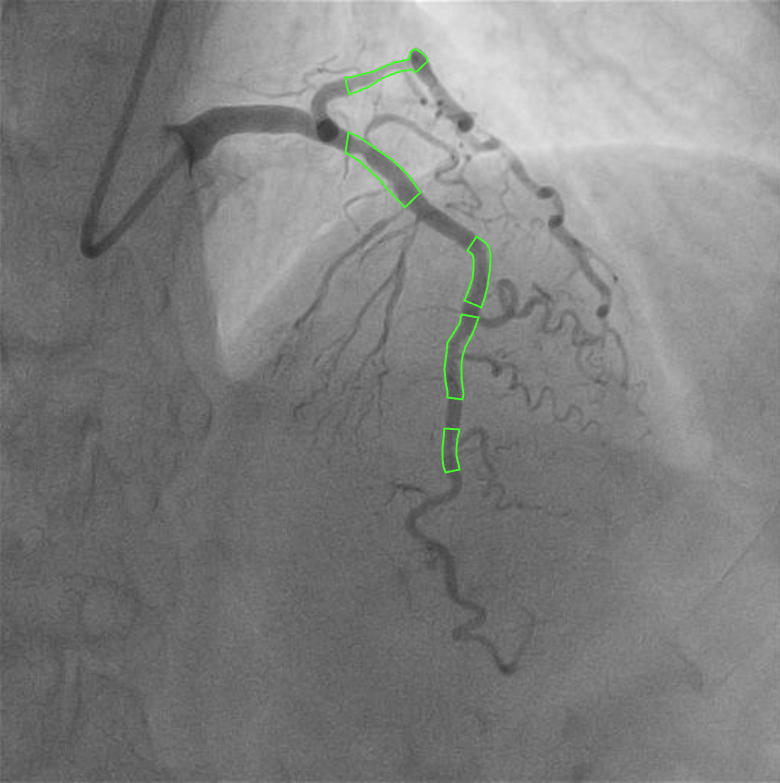}
        \includegraphics[width=\linewidth]{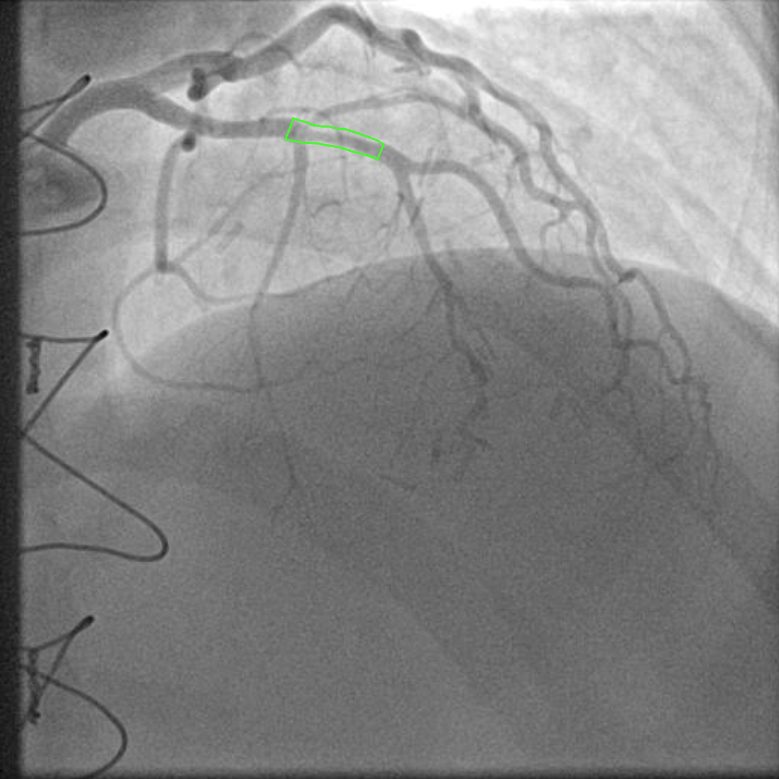}
        \includegraphics[width=\linewidth]{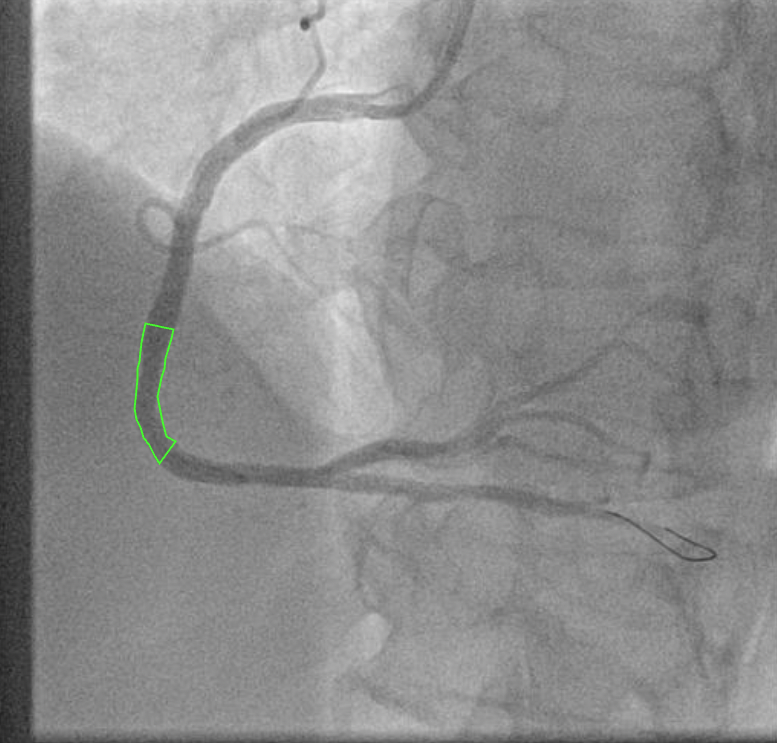}
        \includegraphics[width=\linewidth]{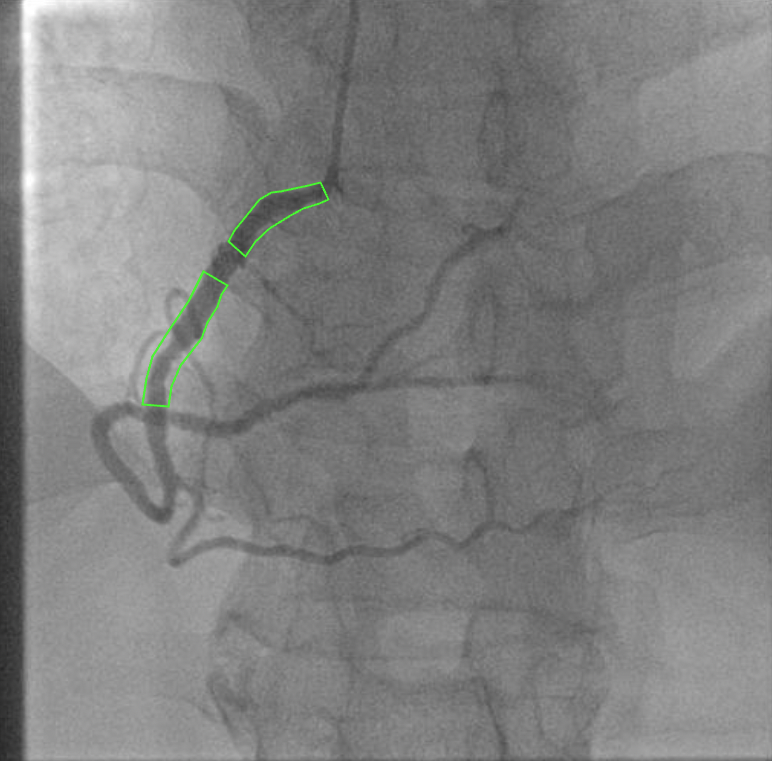}
        \subcaption{Ground truth}
        \label{fig:qual_gt}
    \end{minipage}\hfill
    \begin{minipage}{0.24\linewidth}
        \includegraphics[width=\linewidth]{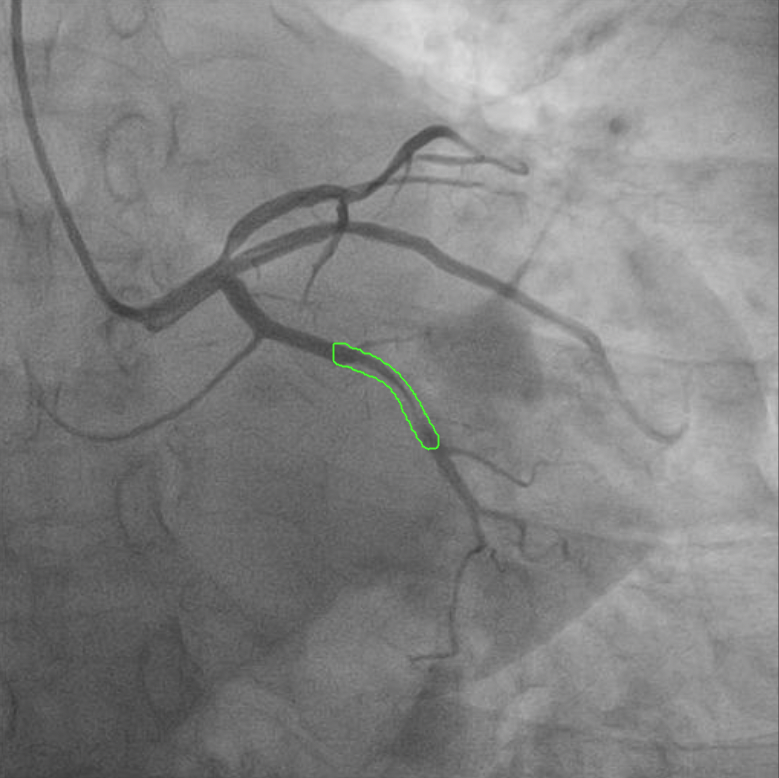}
        \includegraphics[width=\linewidth]{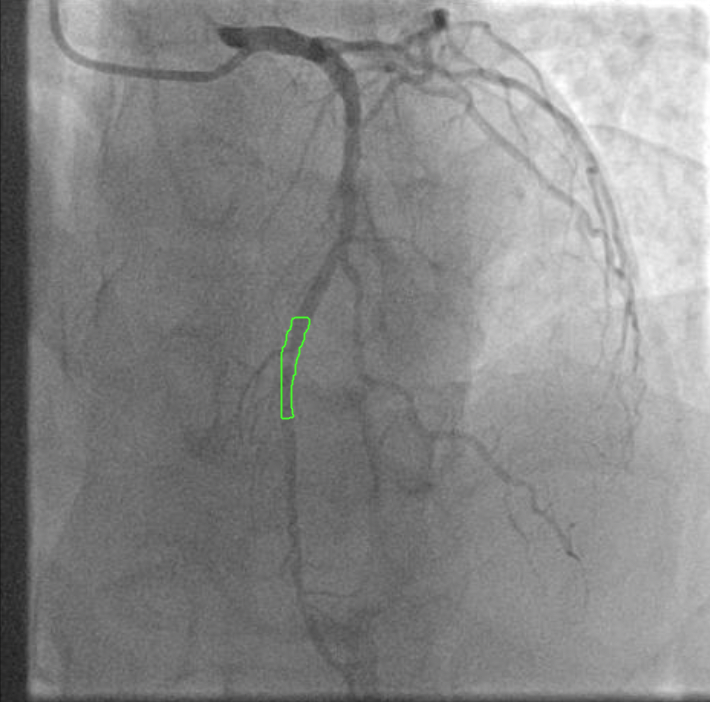}
        \includegraphics[width=\linewidth]{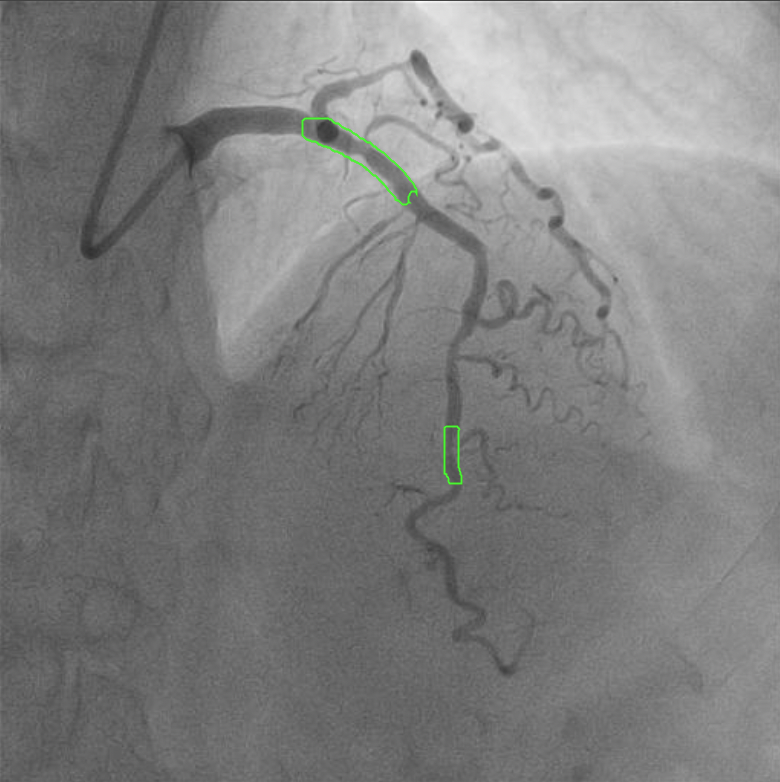}
        \includegraphics[width=\linewidth]{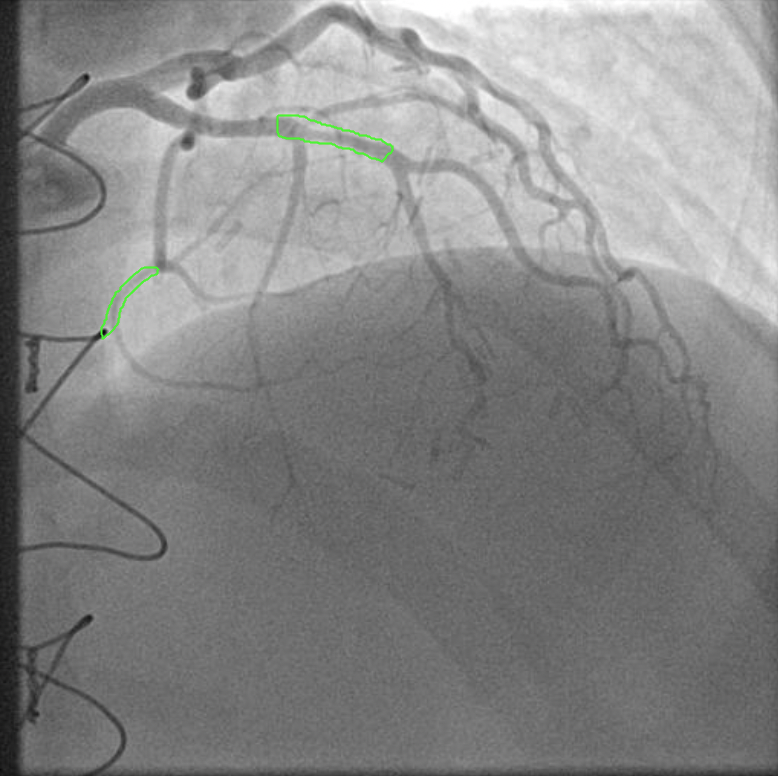}
        \includegraphics[width=\linewidth]{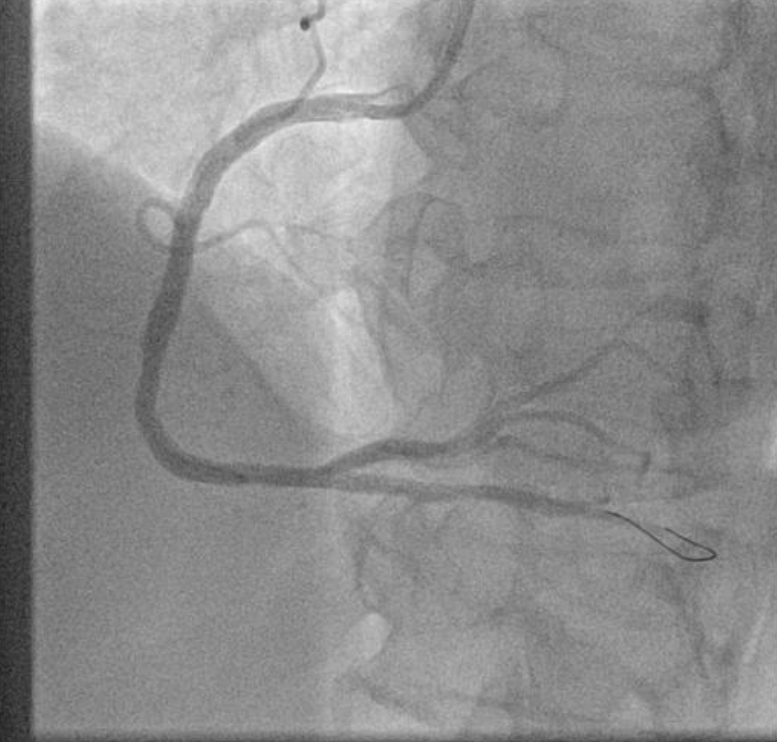}
        \includegraphics[width=\linewidth]{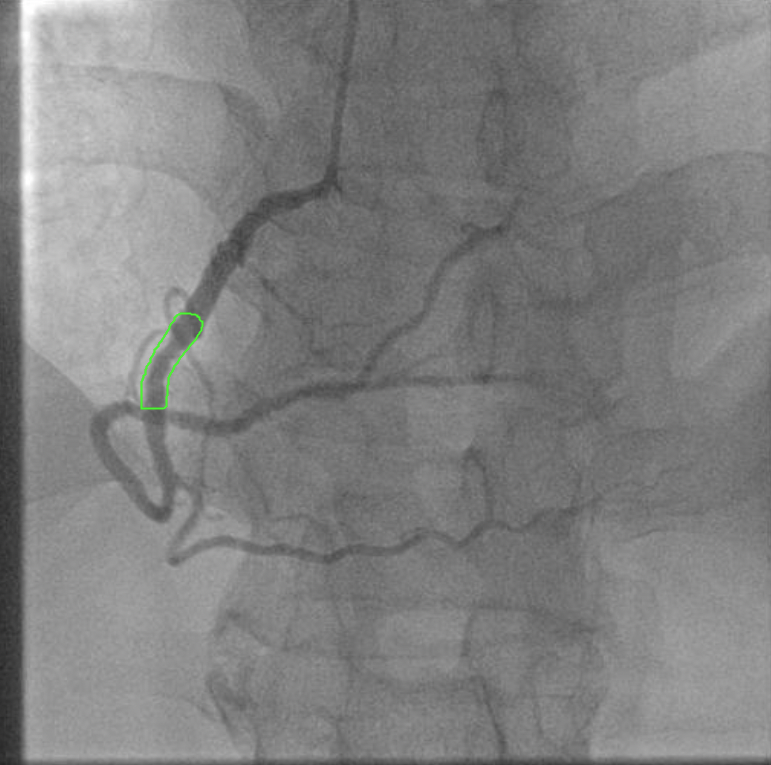}
        \subcaption{Fully-supervised}
        \label{fig:qual_base}
    \end{minipage}\hfill
    \begin{minipage}{0.24\linewidth}
        \includegraphics[width=\linewidth]{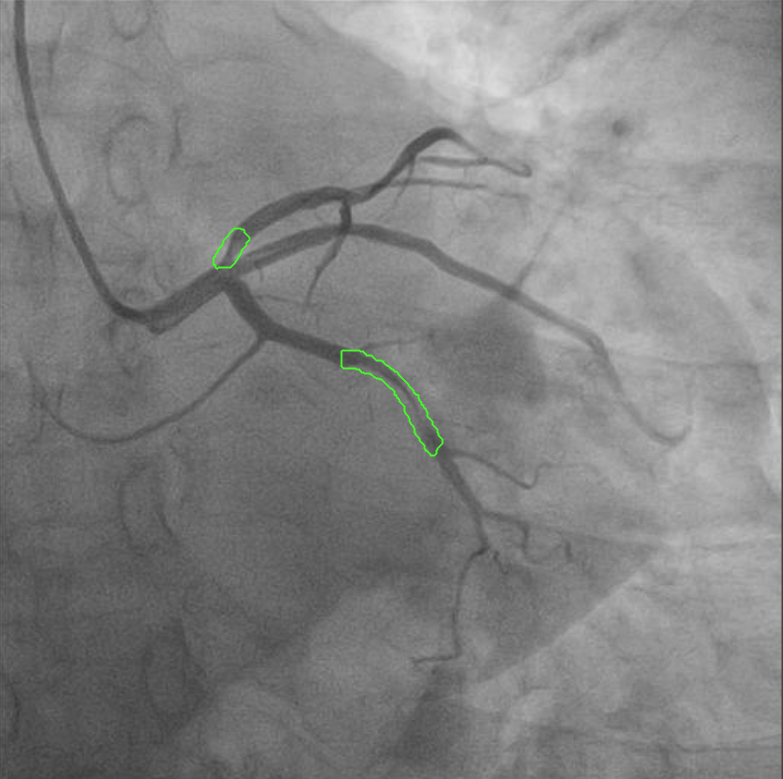}
        \includegraphics[width=\linewidth]{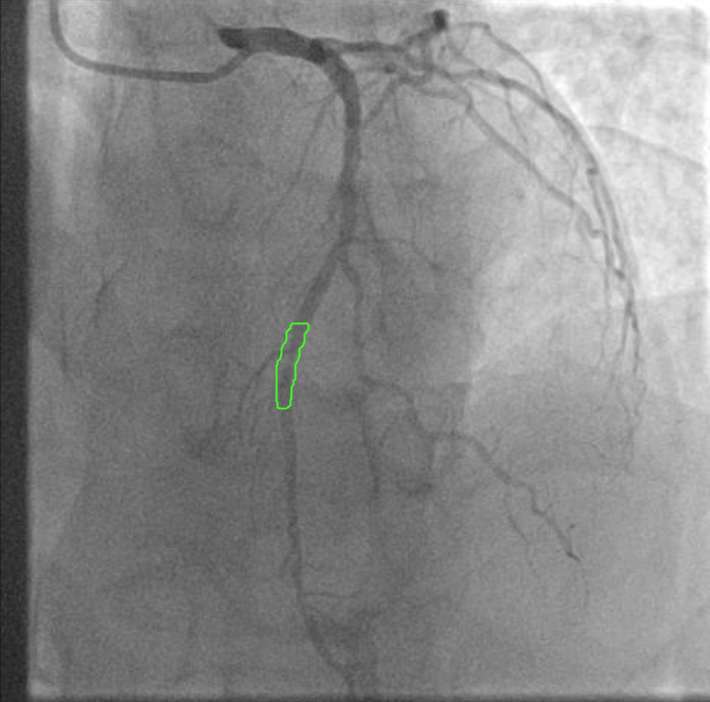}
        \includegraphics[width=\linewidth]{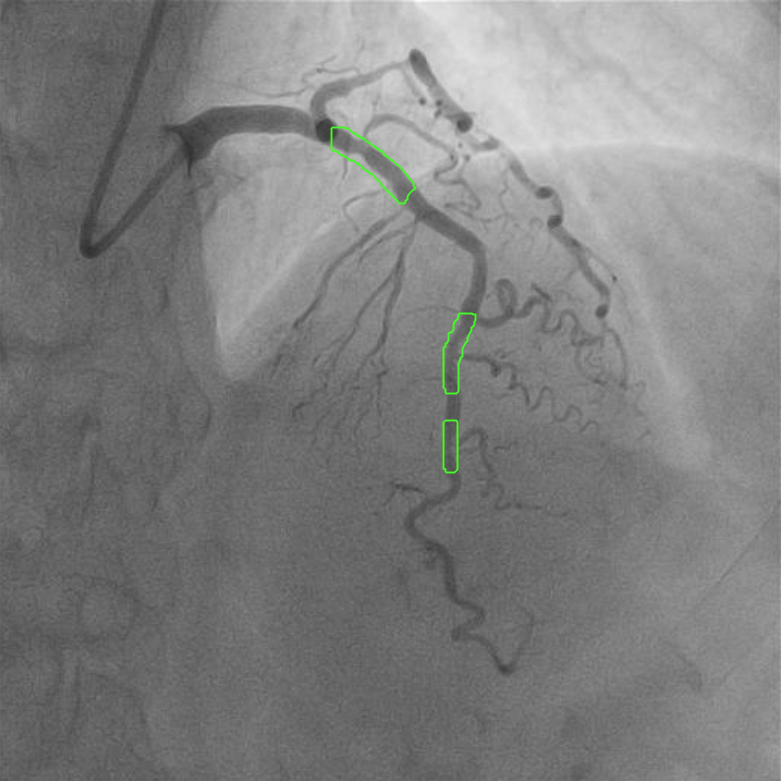}
        \includegraphics[width=\linewidth]{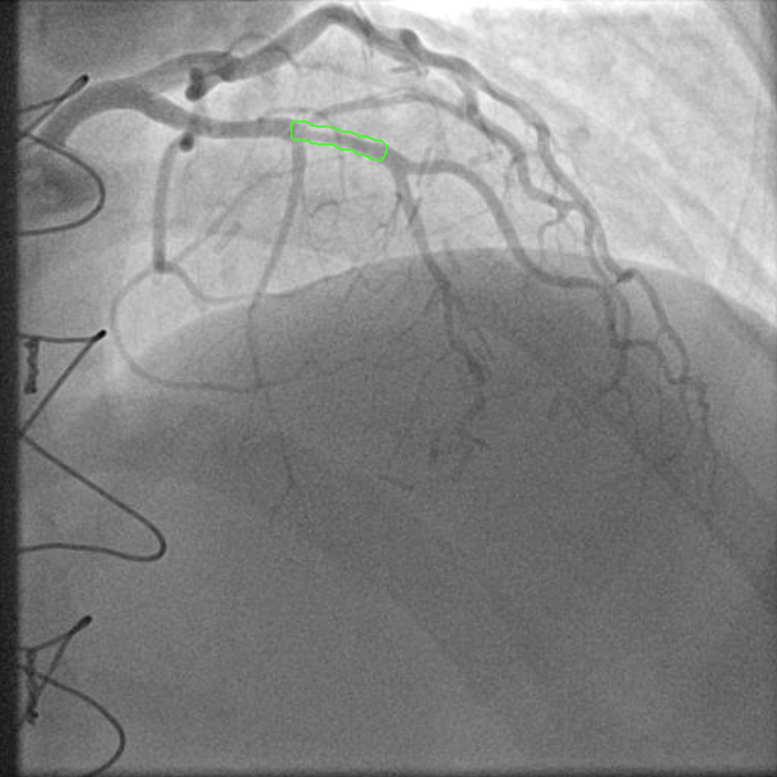}
        \includegraphics[width=\linewidth]{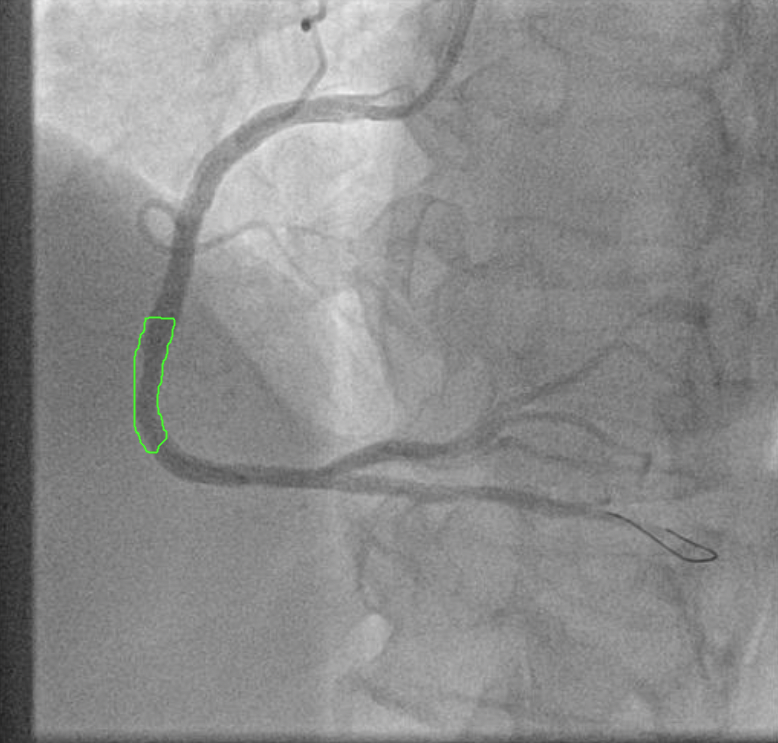}
        \includegraphics[width=\linewidth]{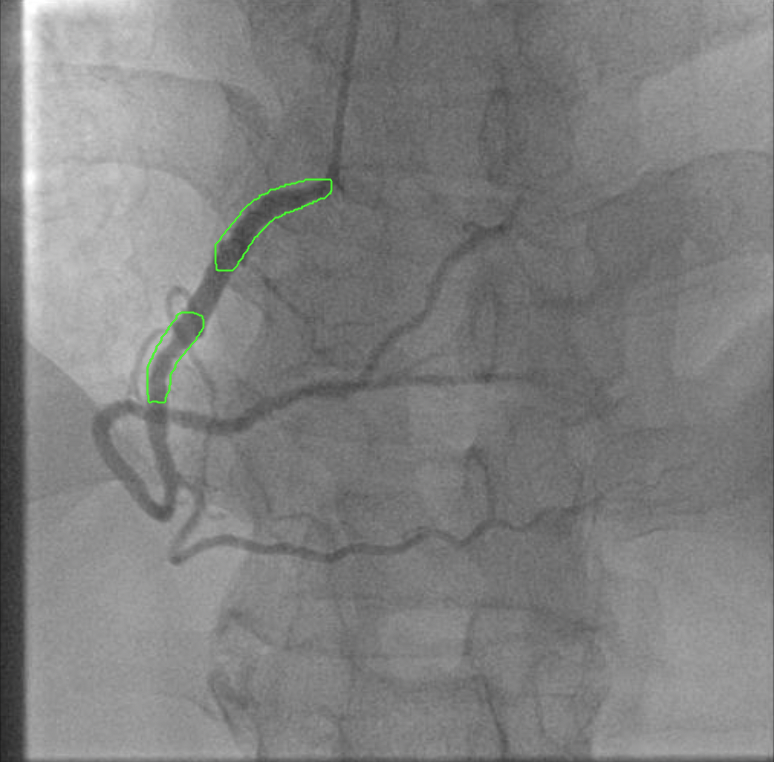}
        \subcaption{Semi-supervised}
        \label{fig:qual_pseudo}
    \end{minipage}\hfill
    
    \label{fig:qualitative_result}
\caption{
Qualitative results of our supervised and semi-supervised methods. Stenosis regions are contoured with light green lines. While quality of stenosis contours between the semi-supervised model and the fully-supervised model is similar, the semi-supervised model identifies more stenotic regions than the fully-supervised model.
}
\end{figure}

\section{Results and Discussion}

\ik{
The ARCADE challenge was structured into two phases, each with its own evaluation and submission criteria. 
In the first phase, where the labels of the valid dataset were not provided, we took a proactive approach. 
To facilitate model assessment, we partitioned the provided training dataset and conducted a 5-fold cross-validation. 
Based on the outcomes of the cross-validation, we trained models and submitted predictions on the images from the valid dataset.
In the final phase of the challenge, the valid dataset with labels was released. 
During this phase, our submitted docker container underwent evaluation using the test dataset. 
To determine the final model for submission in the challenge, we relied on the scores obtained from the valid dataset, ensuring that our selected model exhibited strong performance and robustness in this evaluation phase.

Our final quantitative results for fully-supervised and semi-supervised methods are reported in Table \ref{tab2}. 
On the test dataset, our fully-supervised model achieved a mean F1 score of $0.520$, while our semi-supervised model exhibited an improved mean F1 score of $0.536$. 
It is worth noting that there was an overall improvement in mean F1 scores across other models, including YOLOv8n and YOLOv8s.

Qualitative results comparing fully-supervised and semi-supervised methods are presented in Fig. \ref{fig:qual_img}. 
In terms of the quality of the segmentation outcomes, both the fully-supervised and semi-supervised models exhibited similar characteristics.  
However, in some instances, the fully-supervised model's predictions achieved higher F1 scores than those of the semi-supervised model. 
This discrepancy is likely attributable to the lower quality of the stenotic masks in the pseudo-labels generated for the semi-supervised model.

However, as Fig. \ref{fig:qual_img} illustrates, our semi-supervised model excels in the task of detecting and identifying stenosis regions when compared to the fully-supervised model. 
This advantage arises from the fact that the semi-supervised model was trained on a more diverse dataset, incorporating the additional information from pseudo-labels. 
In the context of the ARCADE challenge, where false positive and false negative instances are counted as 0 when calculating mean F1 scores, the ability to accurately identify every stenosis instance assumes greater importance than refining the quality of the segmentation alone. 

Throughout the course of the challenge, we explored various approaches to enhance our semi-supervised learning strategy.
For instance, we experimented with increasing the number of pseudo-labels by lowering the confidence threshold for predictions. 
While this adjustment did lead to quicker training, it also exacerbated the issue of overfitting, which hindered overall performance. 
We also attempted an alternative strategy where we pretrained the model using the pseudo-label dataset and subsequently fine-tuned it on the original stenosis dataset. 
However, this approach did not yield as high a mean F1 score as the model trained with the combined dataset.
In another attempt to improve performance, we considered adding images without stenosis, with the expectation that it might enhance the model's ability to generalize.
Unfortunately, this approach did not yield the desired results, as both the valid and test datasets contained images with at least one stenosis region.

Ultimately, despite our exploration of various techniques, it was the simplest and most straightforward method that proved to be the most effective in achieving the best performance in the ARCADE challenge.


The nature of challenges often necessitates a different approach compared to conventional research endeavors.
The primary objective in challenges is typically to achieve the highest possible evaluation metrics within time constraints.
While certain experiments showed considerable promise, not all of them were brought to completion. 
For instance, there is the unexplored option of utilizing soft labels for the pseudo-label dataset.
Exploring ways to optimize soft labels for pseudo-labels in future work could offer a promising avenue for enhancing performance.

}

\begin{table}[H]
\renewcommand{\abovecaptionskip}{1pt} 
\renewcommand{\belowcaptionskip}{1pt}
\caption{Data augmentation hyperparameters.}
\label{tab1}
\centering
\begin{tabular}{|c|c|c|}
\hline
Augmentation & Value & Probability\\
\hline
vertical flip    &  -             &  0.5      \\
horizontal flip  &  -             &  0.5      \\
translate        & 0.3            &  uniform  \\
rotation         & 30$^{\circ}$   &  uniform  \\
scale            & 0.5            &  uniform  \\
shear            & 5.0$^{\circ}$  &  uniform  \\
perspective      & 0.001          &  uniform  \\
hue              & 0.015          &  uniform  \\
saturation       & 0.7            &  uniform  \\
value            & 0.4            &  uniform  \\
\hline
\end{tabular}
\end{table}

\begin{table}[H]
\caption{meanF1 scores of fully-supervised and semi-supervised methods in the test dataset.}
\label{tab2}
\centering
\begin{tabular}{|c|c|c|}
\hline
Model & Fully-supervised & Semi-supervised\\
\hline
YOLOv8n  &  0.491  &  0.507\\
YOLOv8s  &  0.515  &  0.520\\
YOLOv8m  &  0.520  &  \textbf{0.536}\\
YOLOv8l  &  0.526  &  0.530\\
\hline
\end{tabular}
\end{table}

\section{Conclusion}

\hw{
In this study, we have proposed and implemented an effective strategy for segmenting cardiovascular stenosis in CAG. 
Our approach commenced with the execution of data augmentation, specifically tailored to reflect the structural characteristics of coronary arteries. 
Subsequently, we employed a pseudo-label-based semi-supervised learning technique, utilizing the data augmented from the initial phase. 
Remarkably, our learning strategy demonstrated top performance in the ARCADE-Stenosis Detection Algorithm. 
This achievement was made possible without relying on an ensemble of multiple models but rather leveraging a straightforward model such as YOLOv8. 
This result underscores our approach's efficiency and effectiveness in addressing complex medical imaging challenges.
}

\bibliographystyle{splncs04}
\typeout{}
\bibliography{main}

\begin{thebibliography}{10}
\providecommand{\url}[1]{\texttt{#1}}
\providecommand{\urlprefix}{URL }
\providecommand{\doi}[1]{https://doi.org/#1}

\bibitem{medical_image_processing_survey}
Aljuaid, A., Anwar, M.: Survey of supervised learning for medical image processing. SN Computer Science  \textbf{3}(4), ~292 (2022)

\bibitem{coronary_angiography_textbook}
Bulwer, B.: Coronary Artery Territories: Second Edition, 2020 (Echocardiography Illustrated). Independently published (2020)

\bibitem{ul_ssl_survey}
Chen, Y., Mancini, M., Zhu, X., Akata, Z.: Semi-supervised and unsupervised deep visual learning: A survey (2022)

\bibitem{ref_breast_tumor}
Cho, S.W., Baek, N.R., Park, K.R.: Deep learning-based multi-stage segmentation method using ultrasound images for breast cancer diagnosis. Journal of King Saud University-Computer and Information Sciences  \textbf{34}(10),  10273--10292 (2022)

\bibitem{imagenet}
Deng, J., Dong, W., Socher, R., Li, L.J., Li, K., Fei-Fei, L.: Imagenet: A large-scale hierarchical image database. In: 2009 IEEE conference on computer vision and pattern recognition. pp. 248--255. Ieee (2009)

\bibitem{gan}
Goodfellow, I.J., Pouget-Abadie, J., Mirza, M., Xu, B., Warde-Farley, D., Ozair, S., Courville, A., Bengio, Y.: Generative adversarial networks (2014)

\bibitem{resnet}
He, K., Zhang, X., Ren, S., Sun, J.: Deep residual learning for image recognition. In: Proceedings of the IEEE conference on computer vision and pattern recognition. pp. 770--778 (2016)

\bibitem{autoencoder}
Liou, C.Y., Cheng, W.C., Liou, J.W., Liou, D.R.: Autoencoder for words. Neurocomputing  \textbf{139},  84--96 (2014)

\bibitem{dataset}
Maxim~Popov, A.A., Zhaksylyk, N., Alkanov, A., Saniyazbekov, A., Aimyshev, T., Ismailov, E., Bulegenov, A., Kolesnikov, A., Kulanbayeva, A., Kuzhukeyev, A., Sakhov, O., Kalzhanov, A., Temenov, N., Fazli1, S.: Arcade: Automatic region-based coronary artery disease diagnostics using x-ray angiography images dataset phase 1 (2023). \doi{10.5281/zenodo.7981244}

\bibitem{aha_guideline_23}
Members, W.C., Virani, S.S., Newby, L.K., Arnold, S.V., Bittner, V., Brewer, L.C., Demeter, S.H., Dixon, D.L., Fearon, W.F., Hess, B., et~al.: 2023 aha/acc/accp/aspc/nla/pcna guideline for the management of patients with chronic coronary disease: a report of the american heart association/american college of cardiology joint committee on clinical practice guidelines. Journal of the American College of Cardiology  \textbf{82}(9),  833--955 (2023)

\bibitem{global_burden_19}
Mensah, G.A., Roth, G.A., Fuster, V.: The global burden of cardiovascular diseases and risk factors: 2020 and beyond (2019)

\bibitem{pseudo_label_ssl}
Min, Z., Ge, Q., Tai, C.: Why the pseudo label based semi-supervised learning algorithm is effective? (2023)

\bibitem{unet}
Ronneberger, O., Fischer, P., Brox, T.: U-net: Convolutional networks for biomedical image segmentation. In: Medical Image Computing and Computer-Assisted Intervention--MICCAI 2015: 18th International Conference, Munich, Germany, October 5-9, 2015, Proceedings, Part III 18. pp. 234--241. Springer (2015)

\bibitem{global_burden_20}
Roth, G.A., Mensah, G.A., Fuster, V.: The global burden of cardiovascular diseases and risks: a compass for global action (2020)

\bibitem{cad_lead_to_16}
Sanchis-Gomar, F., Perez-Quilis, C., Leischik, R., Lucia, A.: Epidemiology of coronary heart disease and acute coronary syndrome. Annals of translational medicine  \textbf{4}(13) (2016)

\bibitem{augmentation_survey}
Shorten, C., Khoshgoftaar, T.M.: A survey on image data augmentation for deep learning. Journal of Big Data  \textbf{6}(1), ~60 (Jul 2019). \doi{10.1186/s40537-019-0197-0}, \url{https://doi.org/10.1186/s40537-019-0197-0}

\bibitem{syntax_origin}
Sianos, G., Morel, M.A., Kappetein, A.P., Morice, M.C., Colombo, A., Dawkins, K., van~den Brand, M., Van~Dyck, N., Russell, M.E., Mohr, F.W., et~al.: The syntax score: an angiographic tool grading the complexity of coronary artery disease. EuroIntervention  \textbf{1}(2),  219--227 (2005)

\bibitem{efficientnet}
Tan, M., Le, Q.V.: Efficientnet: Rethinking model scaling for convolutional neural networks. CoRR  \textbf{abs/1905.11946} (2019), \url{http://arxiv.org/abs/1905.11946}

\bibitem{aha_stats_23}
Tsao, C.W., Aday, A.W., Almarzooq, Z.I., Anderson, C.A., Arora, P., Avery, C.L., Baker-Smith, C.M., Beaton, A.Z., Boehme, A.K., Buxton, A.E., et~al.: Heart disease and stroke statistics—2023 update: a report from the american heart association. Circulation  \textbf{147}(8),  e93--e621 (2023)

\bibitem{global_burden_22}
Vaduganathan, M., Mensah, G.A., Turco, J.V., Fuster, V., Roth, G.A.: The global burden of cardiovascular diseases and risk: a compass for future health (2022)

\bibitem{ssl_survey}
Van~Engelen, J.E., Hoos, H.H.: A survey on semi-supervised learning. Machine learning  \textbf{109}(2),  373--440 (2020)

\bibitem{who_cvd_rpt}
World-Health-Organization: Cardiovascular diseases. \url{https://www.who.int/news-room/fact-sheets/detail/cardiovascular-diseases-(cvds)}, accessed: 2023-09-28

\end{thebibliography}

\end{document}